\documentclass{article}

\usepackage{PRIMEarxiv}

\usepackage[utf8]{inputenc} 
\usepackage[T1]{fontenc}    
\usepackage{hyperref}       
\usepackage{url}            
\usepackage{booktabs}       
\usepackage{amsfonts}       
\usepackage{nicefrac}       
\usepackage{microtype}      
\usepackage{lipsum}
\usepackage{fancyhdr}       
\usepackage{graphicx}       
\graphicspath{{media/}}     
\usepackage{bbm}
\usepackage{amsmath}
\usepackage{amssymb}
\usepackage{multirow}
\usepackage{appendix}

\usepackage[ruled,vlined]{algorithm2e}
\usepackage{xcolor}

\title{HOME: High-Order Mixed-Moment-based Embedding for Representation Learning
}

\author{
  Chuang Niu and Ge Wang \\
  Biomedical Imaging Center \\
  Department of Biomedical Engineering \\
  School of Engineering \\
  Center for Biotechnology and Interdisciplinary Studies \\
  Rensselaer Polytechnic Institute \\
  Troy, New York, USA \\
  \texttt{\{niuc, wangg6\}@rpi.edu} \\
}

\begin{document}
\maketitle

\begin{abstract}
Minimum redundancy among different elements of an embedding in a latent space is a fundamental requirement or major preference in representation learning to capture intrinsic informational structures.
Current self-supervised learning methods minimize a pair-wise covariance matrix to reduce the feature redundancy and produce promising results.
However, such representation features of multiple variables may contain the redundancy among more than two feature variables that cannot be minimized via the pairwise regularization.
Here we propose the High-Order Mixed-Moment-based Embedding (HOME) strategy to reduce the redundancy between any sets of feature variables, which is to our best knowledge the first attempt to utilize high-order statistics/information in this context.
Multivariate mutual information is minimum if and only if multiple variables are mutually independent, which suggests the necessary conditions of factorized mixed moments among multiple variables.
Based on these statistical and information theoretic principles, our general HOME framework is presented for self-supervised representation learning.
Our initial experiments show that a simple version in the form of a three-order HOME scheme already significantly outperforms the current two-order baseline method (i.e., Barlow Twins) in terms of the linear evaluation on representation features.

\end{abstract}


\section{Introduction}

Representation learning that maps high-dimensional data into the semantic features is a fundamental task in computer vision, machine learning, and artificial intelligence \cite{bengio2013representation}.
In particular, self-supervised representation learning (SSRL) on large-scale unlabeled datasets has made great progress and been applied to various applications, such as object detection and segmentation \cite{moco}, deep clustering \cite{niu2021spice}, medical image analysis \cite{tang2022self, niu2022unsupervised}, etc.
To learn meaningful representations without annotations, various pretext tasks were heuristically designed for SSRL, such as denoising auto-encoders \cite{vincent2008extracting}, context auto-encoders \cite{pathak2016context}, cross-channel auto-encoders or colorization \cite{colorization2016, zhang2017split}, masked auto-encoders \cite{he2022masked}, rotation~\cite{rotation2018}, patch ordering~\cite{jigsaw2016, doersch2015unsupervised, chen2021jigsaw}, clustering \cite{caron2018deep, caron2019unsupervised, asano2019self}, and instance discrimination \cite{dosovitskiy2014discriminative, wu2018unsupervised}.
Recently, semantic invariance to predefined transformations of the same instance has been used as a basic pretext task in various SSRL methods \cite{simclr, moco, grill2020bootstrap, Chen_2021_CVPR, caron2020unsupervised} due to its effectiveness and efficiency.
To avoid trivial solutions (e.g., the features of all samples are a constant vector), these methods use various special techniques, such as large batches or a memory bank \cite{wu2018unsupervised, misra2020self, simclr}, momentum updating \cite{moco}, asymmetry network architecture with additional predictor head and stop gradients \cite{grill2020bootstrap, Chen_2021_CVPR}.
In another direction, W-MSE \cite{ermolov2021whitening}, Barlow~Twins \cite{zbontar2021barlow}, and VICReg \cite{bardes2021vicreg} drive covariance matrices towards the identity matrix to minimize the pairwise correlation, explicitly avoiding trivial solutions without requiring any asymmetric constraint on network architectures nor training process.

Based on the general characteristics of the expected embedding features instead of ad-hoc techniques for SSRL, here we propose a principled approach for self-learning.
Clearly, a naturally desired property is that semantically similar samples have similar embedding features.
This can be approximately achieved by the common pretext task of transform invariance that different transformations of the same instance should have the same embedding features. The transformation is randomly performed according to a predefined transform distribution, such as random cropping, horizontal flip, color jittering, grayscale, Gaussian blur, and solarization, assuming that these transformations will not affect the semantic meanings of the original instance.
This study highlights the importance of the minimum redundancy among all feature variables; that is, the mutual information between any sets of variables is minimized.
With this minimum redundancy constraint, learned features are enriched, concentrated, and decomposed to be informative. 
Although existing studies \cite{ermolov2021whitening, zbontar2021barlow, bardes2021vicreg} proposed to minimize the pairwise correlation by enforcing the off-diagonal elements of a covariance matrix to be zero, the minimum redundancy among multiple feature variables has not been utilized so far.
Theoretically, multivariate mutual information \cite{mcgill1954multivariate} or total correlation \cite{watanabe1960information} is minimum if and only if a set of multiple variables are mutually independent, note that pairwise independence cannot ensure mutually independence \cite{gallager2013stochastic}.
Furthermore, the mutual independence means that the mixed moment of multiple feature variables can be factorized into the multiplication of their individual moments.
Based on this important observation, we advocate a general framework for High-Order Mixed-Moment-based Embedding (HOME) to empower self-supervised representation learning.

The rest of this paper is organized as follows. In the second section, we describe our general methodology. In the third section, we report our initial experimental results on CIFAR-10. As the first step, a three-order HOME was instantiated and evaluated. The results on CIFAR-10 show that the HOME scheme has achieved a significantly better performance on representation learning ($4.1\%$) than the two-order baseline, i.e., Barlow Twins \cite{zbontar2021barlow} on the CIFAR-10 dataset, in terms of the commonly used linear classification evaluation on fixed representation features. In the last section, we discuss relevant issues and conclude the paper.

\section{Methodology}

\subsection{General Self-Supervised Framework}
\begin{figure*}[h]
    \centering
    \includegraphics[width=1\textwidth]{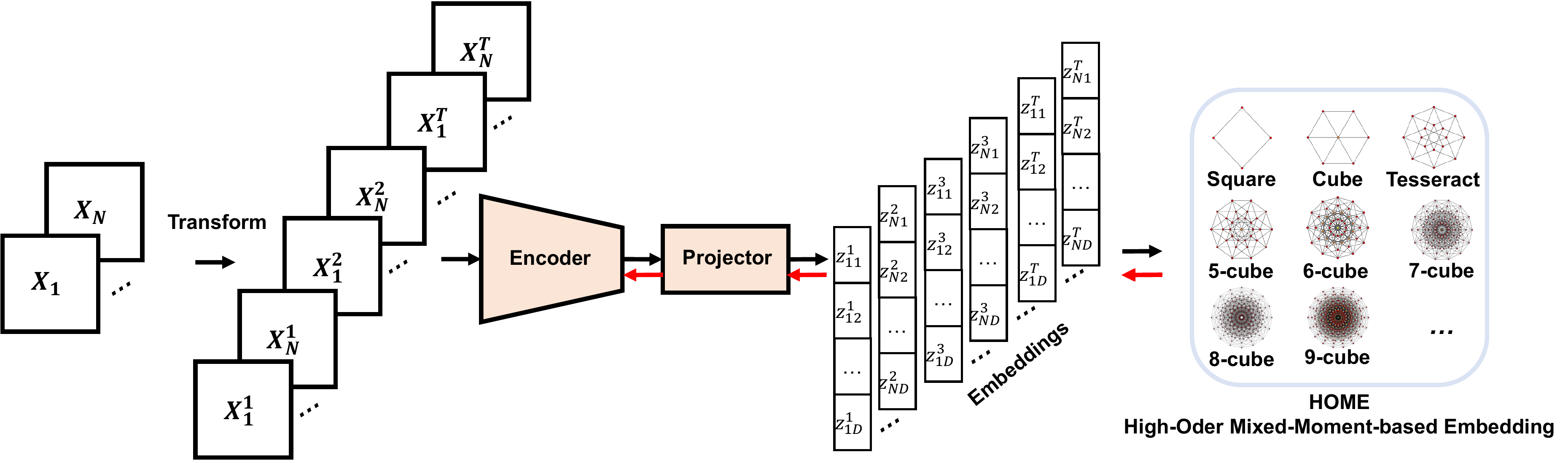}
    \caption{HOME framework for self-supervised representation learning. The black arrows denote the forward data flow, and the red arrows denote the backward gradient flow. Here the high-order constraints among multiple features variables are illustrated with hypercubes.}
    \label{fig_ssl}
\end{figure*}

We aim to train a neural network to  extract  meaningful features on an unlabeled dataset in a self-supervised learning manner.
Our general SSRL framework HOME is shown in Fig. \ref{fig_ssl}, which is formulated into a single branch in the training stage.
In each training iteration, a batch of training samples $\{x_n\}_{n=1}^N$ are first randomly transformed to $T$ distorted versions $\{\{x^t_n\}_{n=1}^N\}_{t=1}^T$. In practice, $T$ is usually set to 2 due to the memory and computational cost although some studies \cite{caron2020unsupervised} achieved better results using more than two transformations.
Then, all transformed images are forwarded to a single model, including an encoder mapping data to the representation features and a projector mapping the representation features to the embedding features; i.e., $z^t_n = G(F(x^t_n; \theta_F); \theta_G) \in R^D$, where $F(\cdot, \theta_F)$ denotes the encoder function with a vector of parameters $\theta_F$, $G(\cdot, \theta_G)$ denotes the projector function with another vector of parameters $\theta_G$, and $D$ is the dimension of embedding features.
This study focuses on learning the generally expected properties of meaningful embedding features, requiring any special constraints on neither the network architecture nor the optimization process.

\subsection{High-Order Mixed-Moment-based Embedding}
Two properties are expected for meaningful embedding features, i.e., invariance to random transformations and minimum total correlation among all feature variables.
The invariance is to drive semantically similar samples close to each other in the embedding space, which has been well established as a basic pretext task in various SSRL methods.
Here we underline the reduction of the total correlation among all feature variables so that informative features can be learned into a compact vector, very similar to coordinates of a point in a Cartesian coordinate system
The total correlation \cite{watanabe1960information, van2011two} of random variables $z_1, z_2, \cdots, z_D$ is defined as follows:
\begin{equation}
    I(Z_1, Z_2, \cdots, Z_D) = \int_{z_1} \int_{z_2} \cdots \int_{z_D} P(z_1, z_2, \cdots, z_D) \log \frac{P(z_1, z_2, \cdots, z_D)}{P(z_1)P(z_2)\cdots P(z_D)} dz_1 dz_2 \cdots dz_D.
\end{equation}
$I(Z_1, Z_2, \cdots, Z_D)$ measures the amount of information shared among multiple random variables. It can be seen that $I(Z_1, Z_2, \cdots, Z_D)$ is minimized if and only if the joint probability density distribution (PDF) can be factorized into their individual PDFs; i.e., $P(z_1, z_2, \cdots, z_D) = P(z_1)P(z_2)\cdots P(z_D)$, which means that all variables are mutually independent.
Note that pairwise independence cannot ensure the mutual independence of the entire set of random variables \cite{gallager2013stochastic}, which means that even if the mutual information between every two variables are zero, the multivariate mutual information may still have not been minimized.
Therefore, to systematically reduce the informational redundancy among all feature variables, the total correlation should be minimized.
Nevertheless, it is generally difficult to estimate the probability distribution of continuous variables so that $I(Z_1, Z_2, \cdots, Z_D)$ cannot be directly minimized.
Furthermore, if all variables are mutually independent, then for every $K$ variables, $K \le D$, and for any $K$ indices  $1 \le d_1 \le \cdots \le d_K \le D$, we have
\begin{equation}
    \begin{split}
        E[\prod_{k=1}^{K} Z_{d_k}] &= \int_{z_{d_1}} \int_{z_{d_2}} \cdots \int_{z_{d_k}} P(z_{d_1}, z_{d_2}, \cdots, z_{d_k})dz_{d_1} dz_{d_2} dz_{d_k} \\
        &=  \int_{z_{d_1}}P(z_{d_1})dz_{d_1}  \int_{z_{d_2}}P(z_{d_2})dz_{d_2} \cdots  \int_{z_{d_k}} P(z_{d_k}) dz_{d_k} \\
        &= \prod_{k=1}^{K}E[Z_{d_k}],
    \end{split}
    \label{eq_fact}
\end{equation}
which means that the mixed moments, $E[\prod_{i=1}^{K} Z_{d_k}]$, among $K$ variables can be factorized to the multiplication of their individual expectations. These expected values can be estimated as the means of observed samples.
When $K=2$, the general mixed moment is degraded to the pairwise correlation used in \cite{ermolov2021whitening, zbontar2021barlow, bardes2021vicreg}.
If and only if the joint distribution $P(z_{d_1}, z_{d_2}, \cdots, z_{d_k})$ is a multivariate normal distribution, the pairwise zero correlation is equivalent to the mutual independence or minimum total correlation.
Unfortunately, the joint normal distribution among all features variables cannot be ensured in practice.
Hence, the necessary conditions of factorizable mixed moments in Eq. (\ref{eq_fact}) must be satisfied to drive the total correlation towards zero.

Based on the above analysis, we presented a HOME loss as
\begin{equation}
    L = \frac{1}{D} \sum_{d=1}^D \frac{2}{T(T-1)} \sum_{i,j=1, i\ne j}^{T} (1 - \sum_{n=1}^N \hat{z}_{nd}^i \hat{z}_{nd}^j)^2 + \lambda \frac{1}{T} \sum_{t=1}^T \frac{1}{M} \sum_{K=2}^D \sum_{d1, d2, \cdots, d_K} (\frac{1}{N} \sum_{n=1}^N \prod_{i=1}^{K} \hat{z}^t_{n d_k})^2,
    \label{eq_loss}
\end{equation}
where all feature variables $z_d$ are normalized with the zero mean and the unit standard deviation, denoted by $\hat{z}_d^t$ as follows:
\begin{equation}
    \hat{z}^t_{nd} = \frac{z^t_{nd} - \frac{1}{N}\sum_{n=1}^N z^t_{nd}}{\sqrt{\sum_{n=1}^N (z^t_{nd} - \frac{1}{N}\sum_{n=1}^N z^t_{nd})^2}}.
    \label{eq_norm}
\end{equation}
The first term in Eq. (\ref{eq_loss}) is to enforce the embedding features from different transformations of the same instance to be the same, which is a mutli-view variant of the one used in \cite{zbontar2021barlow}.
The second term is our novel contribution that constrains the empirical mixed moments to be subject to Eq. (\ref{eq_fact}); i.e., $E[\prod_{k=1}^{K} \hat{Z}_{d_k}]=\prod_{k=1}^{K}E[\hat{Z}_{d_k}]=0$, as $\forall d_k, E[\hat{Z}_{d_k}]=0$ after normalization in Eq. (\ref{eq_norm}), and
$M=\sum_{K=2}^D \frac{D!}{(D-K)!K!}$ is the total number of combinations for all orders of moments, where $K$ denotes the order of moments. In this study, we simply set $\lambda=1$.

\begin{figure*}[h]
    \centering
    \includegraphics[width=1\textwidth]{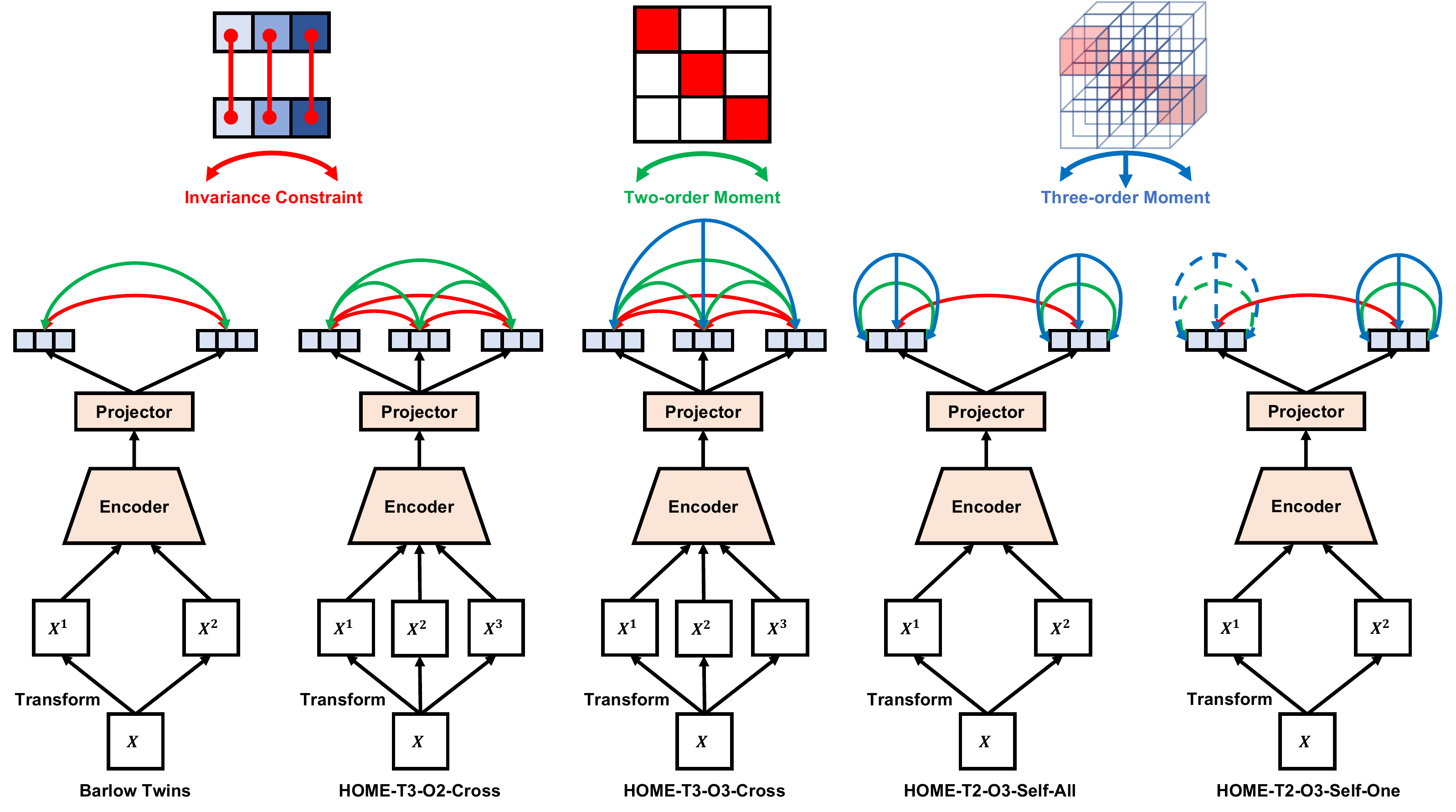}
    \caption{Different variant models. The red square or cube elements are ignored in computing the HOME loss. The dashed lines denote the corresponding transformation is not constrained with self-mixed-moments.}
    \label{fig_compare}
\end{figure*}
\subsection{Three-Order HOME}
\label{sec_implement}
Based on the general HOME framework for SSRL, a three-order HOME self-supervised learning method was instantiated; i.e., $K\in \{2, 3\}$.
In this initial study, we used Barlow Twins \cite{zbontar2021barlow} with the two-order covariance constraint as the baseline.
During the training process of Barlow Twins, two random transformations were forwarded to twin branches and the cross-covariance was calculated.
Thus, the difference of the two-order HOME with $T=2$ from Barlow Twins is that self-covariance is imposed to two transformations separately.
To evaluate the effect of our high-order constraint, different variants of two- and three-order HOME were built and evaluated.
Specifically, the compared variants include:
\begin{itemize}
    \item Barlow Twins: Original Barlow Twins using two transformations and the cross-covariance constraint.
    \item HOME-T3-O2-Cross: Two-order HOME using three transformations and the cross-covariance constraint, where three cross-covariance matrices between every two transformations are calculated.
    \item HOME-T3-O3-Cross: Three-order HOME using three transformations and cross-mixed-moments, where three cross-covariance matrices between every two transformations and a three-order cross-mixed-moment tensor are calculated.
    \item HOME-T2-O3-Self-All: Three-order HOME using two transformations and self-mixed-moments, where two self-covariance matrices and two three-order cross-mixed-moment tensors are separately imposed on two transformations.
    \item HOME-T2-O3-Self-One: Three-order HOME using two transformations and self-mixed-moments, where one self-covariance matrix and one three-order self-mixed-moment tensor are imposed on one of the two transformations randomly.
\end{itemize}
The above variant models are summarized in Fig. \ref{fig_compare}.

\subsection{Implementation Details}

As the first example, the CIFAR-10 dataset was used to evaluate all the model variants.
The commonly used ResNet18 was used as the feature encoder and the three-layer MPL with the dimension 1024 for each layer was used as the projector, which implies the embedding feature dimension $D=1024$.
The SGD optimizer was used with the momentum 0.9 and weight decay rate 0.0005.
We used the cosine decay schedule from 0 with 10 warmup epochs towards the final value 0.002. The base learning rate was set to 0.5. The batch size was set to 512. All models were optimized with 800 epochs on a single Tesla V100 GPU.

\section{Experimental Results}

\begin{table}[htp]
  \renewcommand{\arraystretch}{1.5}
  \renewcommand\tabcolsep{8pt}
 \caption{\textbf{Comparison of different methods on CIFAR-10 with linear probing}. Top-1 and Top-5 accuracies (in \%) of ResNet18 are reported. The best two Top-1 accuracies are highlighted in \textbf{bold}.}
  \centering
  \begin{tabular}{c|cc}
     Methods         & Top-1 & Top-5  \\
     \hline
     MoCo       (2020)             &   90.0 &  -  \\
     SimCLR      (2020)           &   87.5 & - \\
     SimSiam      (2020)           &   90.6 & - \\
     BYOL         (2020)           &   \textbf{91.1} & - \\
     SwAV         (2020)           &   88.1 & - \\
     DINO          (2021)           &   88.4 & - \\
     NNCLR       (2021)           &   89.6 & -  \\
     DCL         (2022)           &   87.3 & -  \\
     \hline
     Barlow Twins (2021)           &   87.1 & 99.2 \\
     HOME-T3-O2-Cross       (ours)           &   87.3 & 99.5 \\
     HOME-T3-O3-Cross       (ours)           &   \textbf{91.1} & 99.7 \\
     HOME-T2-O3-Self-All       (ours)           &   \textbf{91.2} & 99.7 \\
     HOME-T2-O3-Self-One       (ours)           &  \textbf{91.2} & 99.7 \\
    \hline
  \end{tabular}
  \label{tab:compare}
\end{table}

The results obtained using different methods on CIFAR-10 are summarized in Table \ref{tab:compare}, where the numbers in the first block were reproduced by Lightly\footnote{https://github.com/lightly-ai/lightly}, and the data in the second block were generated in this study.
Note that the training settings are basically the same between this study and Lightly, including the all models were optimized with the SGD optimizer, batch size of 512, and 800 training epochs. 
The commonly used linear probing was used to evaluate the representation learning performance of different methods.
Specifically, after the self-supervised training, a linear classifier was stacked onto the encoder network with the frozen parameters while the projector was disregarded.
Without using any special constraints, such as asymmetric network structures, momentum updating, memory bank, stop gradient, etc, HOME achieved the state-of-the-art results on the CIFAR-10 dataset.
Significantly, the three-order HOME outperforms the two-order baseline; i.e., Barlow Twins, by a $\sim$4\% in terms of Top-1 accuracy.
As discussed in Section \ref{sec_implement}, we assume that the cross-mixed-moment is equivalent to the self-mixed-moment as the embedding features of different transformations tend to be the same, which is demonstrated by our results.
Furthermore, we do not have to impose the empirical constraints on all transformations, since randomly selecting one seems sufficient to yield the equivalent results, which helps save the computational cost.

\section{Discussions and Conclusion}
The HOME loss function defined in Eq. (\ref{eq_loss}) indicates that with incorporation of high-order mixed moment orders, the self-learning results can be significantly improved, while the computational cost will be substantially increased as well. In this initial investigation, we only used three-order of moments on a relatively small dataset, and thus the HOME loss can be fully computed. However, it may be not feasible to include many moments for optimizing representation learning models on the large-scale datasets, unless more efficient algorithms can be designed or more powerful computing platforms are used.
At this stage, an immediate solution is to randomly select part of high-order elements in each training iteration to fit the limitations of computers. We will report more results on larger datasets in the near future.

In conclusion, this study presents a High-Order Mixed-Moment-based Embedding (HOME) approach for representation learning.
HOME, as the general self-supervised learning framework, is promising to reduce the total correlation among all feature variables, making the features rich and compact.
Without using any ad-hoc techniques, a three-order HOME has already achieved the state-of-the-art results on the CIFAR-10 dataset, and outperformed the two-order baseline; i.e., Barlow Twins, by a large margin (4.1\%) in terms of the linear classification.
HOME is effective to learn the generally expected properties of representation features, and should have a major impact on the deep learning field after it is adapted to refined versions and applied to various tasks in different domains.

\bibliographystyle{unsrt}  
\bibliography{references}

\begin{thebibliography}{10}

\bibitem{bengio2013representation}
Yoshua Bengio, Aaron Courville, and Pascal Vincent.
\newblock Representation learning: A review and new perspectives.
\newblock {\em IEEE transactions on pattern analysis and machine intelligence},
  35(8):1798--1828, 2013.

\bibitem{moco}
Kaiming He, Haoqi Fan, Yuxin Wu, Saining Xie, and Ross Girshick.
\newblock Momentum contrast for unsupervised visual representation learning.
\newblock In {\em Proceedings of the IEEE/CVF Conference on Computer Vision and
  Pattern Recognition (CVPR)}, June 2020.

\bibitem{niu2021spice}
Chuang Niu and Ge~Wang.
\newblock Spice: Semantic pseudo-labeling for image clustering, 2021.

\bibitem{tang2022self}
Yucheng Tang, Dong Yang, Wenqi Li, Holger~R Roth, Bennett Landman, Daguang Xu,
  Vishwesh Nath, and Ali Hatamizadeh.
\newblock Self-supervised pre-training of swin transformers for 3d medical
  image analysis.
\newblock In {\em Proceedings of the IEEE/CVF Conference on Computer Vision and
  Pattern Recognition}, pages 20730--20740, 2022.

\bibitem{niu2022unsupervised}
Chuang Niu and Ge~Wang.
\newblock Unsupervised contrastive learning based transformer for lung nodule
  detection.
\newblock {\em arXiv preprint arXiv:2205.00122}, 2022.

\bibitem{vincent2008extracting}
Pascal Vincent, Hugo Larochelle, Yoshua Bengio, and Pierre-Antoine Manzagol.
\newblock Extracting and composing robust features with denoising autoencoders.
\newblock In {\em Proceedings of the 25th international conference on Machine
  learning}, pages 1096--1103, 2008.

\bibitem{pathak2016context}
Deepak Pathak, Philipp Krahenbuhl, Jeff Donahue, Trevor Darrell, and Alexei~A
  Efros.
\newblock Context encoders: Feature learning by inpainting.
\newblock In {\em Proceedings of the IEEE conference on computer vision and
  pattern recognition}, pages 2536--2544, 2016.

\bibitem{colorization2016}
Richard Zhang, Phillip Isola, and Alexei~A. Efros.
\newblock Colorful image colorization.
\newblock In {\em ECCV}, volume 9907, pages 649--666. Springer, 2016.

\bibitem{zhang2017split}
Richard Zhang, Phillip Isola, and Alexei~A Efros.
\newblock Split-brain autoencoders: Unsupervised learning by cross-channel
  prediction.
\newblock In {\em Proceedings of the IEEE conference on computer vision and
  pattern recognition}, pages 1058--1067, 2017.

\bibitem{he2022masked}
Kaiming He, Xinlei Chen, Saining Xie, Yanghao Li, Piotr Doll{\'a}r, and Ross
  Girshick.
\newblock Masked autoencoders are scalable vision learners.
\newblock In {\em Proceedings of the IEEE/CVF Conference on Computer Vision and
  Pattern Recognition}, pages 16000--16009, 2022.

\bibitem{rotation2018}
Spyros Gidaris, Praveer Singh, and Nikos Komodakis.
\newblock Unsupervised representation learning by predicting image rotations.
\newblock In {\em ICLR}, 2018.

\bibitem{jigsaw2016}
Mehdi Noroozi and Paolo Favaro.
\newblock Unsupervised learning of visual representations by solving jigsaw
  puzzles.
\newblock In {\em ECCV}, pages 69--84, 2016.

\bibitem{doersch2015unsupervised}
Carl Doersch, Abhinav Gupta, and Alexei~A Efros.
\newblock Unsupervised visual representation learning by context prediction.
\newblock In {\em Proceedings of the IEEE international conference on computer
  vision}, pages 1422--1430, 2015.

\bibitem{chen2021jigsaw}
Pengguang Chen, Shu Liu, and Jiaya Jia.
\newblock Jigsaw clustering for unsupervised visual representation learning.
\newblock In {\em Proceedings of the IEEE/CVF Conference on Computer Vision and
  Pattern Recognition}, pages 11526--11535, 2021.

\bibitem{caron2018deep}
Mathilde Caron, Piotr Bojanowski, Armand Joulin, and Matthijs Douze.
\newblock Deep clustering for unsupervised learning of visual features.
\newblock In {\em Proceedings of the European conference on computer vision
  (ECCV)}, pages 132--149, 2018.

\bibitem{caron2019unsupervised}
Mathilde Caron, Piotr Bojanowski, Julien Mairal, and Armand Joulin.
\newblock Unsupervised pre-training of image features on non-curated data.
\newblock In {\em Proceedings of the IEEE/CVF International Conference on
  Computer Vision}, pages 2959--2968, 2019.

\bibitem{asano2019self}
Yuki~Markus Asano, Christian Rupprecht, and Andrea Vedaldi.
\newblock Self-labelling via simultaneous clustering and representation
  learning.
\newblock {\em arXiv preprint arXiv:1911.05371}, 2019.

\bibitem{dosovitskiy2014discriminative}
Alexey Dosovitskiy, Jost~Tobias Springenberg, Martin Riedmiller, and Thomas
  Brox.
\newblock Discriminative unsupervised feature learning with convolutional
  neural networks.
\newblock {\em Advances in neural information processing systems}, 27, 2014.

\bibitem{wu2018unsupervised}
Zhirong Wu, Yuanjun Xiong, Stella~X Yu, and Dahua Lin.
\newblock Unsupervised feature learning via non-parametric instance
  discrimination.
\newblock In {\em Proceedings of the IEEE conference on computer vision and
  pattern recognition}, pages 3733--3742, 2018.

\bibitem{simclr}
Ting Chen, Simon Kornblith, Mohammad Norouzi, and Geoffrey Hinton.
\newblock A simple framework for contrastive learning of visual
  representations.
\newblock In {\em Proceedings of the 37th International Conference on Machine
  Learning}, volume 119, pages 1597--1607, 2020.

\bibitem{grill2020bootstrap}
Jean-Bastien Grill, Florian Strub, Florent Altch{\'e}, Corentin Tallec, Pierre
  Richemond, Elena Buchatskaya, Carl Doersch, Bernardo Avila~Pires, Zhaohan
  Guo, Mohammad Gheshlaghi~Azar, et~al.
\newblock Bootstrap your own latent-a new approach to self-supervised learning.
\newblock {\em Advances in Neural Information Processing Systems},
  33:21271--21284, 2020.

\bibitem{Chen_2021_CVPR}
Xinlei Chen and Kaiming He.
\newblock Exploring simple siamese representation learning.
\newblock In {\em Proceedings of the IEEE/CVF Conference on Computer Vision and
  Pattern Recognition (CVPR)}, pages 15750--15758, June 2021.

\bibitem{caron2020unsupervised}
Mathilde Caron, Ishan Misra, Julien Mairal, Priya Goyal, Piotr Bojanowski, and
  Armand Joulin.
\newblock Unsupervised learning of visual features by contrasting cluster
  assignments.
\newblock {\em Advances in Neural Information Processing Systems},
  33:9912--9924, 2020.

\bibitem{misra2020self}
Ishan Misra and Laurens van~der Maaten.
\newblock Self-supervised learning of pretext-invariant representations.
\newblock In {\em Proceedings of the IEEE/CVF Conference on Computer Vision and
  Pattern Recognition}, pages 6707--6717, 2020.

\bibitem{ermolov2021whitening}
Aleksandr Ermolov, Aliaksandr Siarohin, Enver Sangineto, and Nicu Sebe.
\newblock Whitening for self-supervised representation learning.
\newblock In {\em International Conference on Machine Learning}, pages
  3015--3024. PMLR, 2021.

\bibitem{zbontar2021barlow}
Jure Zbontar, Li~Jing, Ishan Misra, Yann LeCun, and St{\'e}phane Deny.
\newblock Barlow twins: Self-supervised learning via redundancy reduction.
\newblock In {\em International Conference on Machine Learning}, pages
  12310--12320. PMLR, 2021.

\bibitem{bardes2021vicreg}
Adrien Bardes, Jean Ponce, and Yann LeCun.
\newblock Vicreg: Variance-invariance-covariance regularization for
  self-supervised learning.
\newblock {\em arXiv preprint arXiv:2105.04906}, 2021.

\bibitem{mcgill1954multivariate}
William McGill.
\newblock Multivariate information transmission.
\newblock {\em Transactions of the IRE Professional Group on Information
  Theory}, 4(4):93--111, 1954.

\bibitem{watanabe1960information}
Satosi Watanabe.
\newblock Information theoretical analysis of multivariate correlation.
\newblock {\em IBM Journal of research and development}, 4(1):66--82, 1960.

\bibitem{gallager2013stochastic}
Robert~G Gallager.
\newblock {\em Stochastic processes: theory for applications}.
\newblock Cambridge University Press, 2013.

\bibitem{van2011two}
Tim Van~de Cruys.
\newblock Two multivariate generalizations of pointwise mutual information.
\newblock In {\em Proceedings of the Workshop on Distributional Semantics and
  Compositionality}, pages 16--20, 2011.

\end{thebibliography}

\end{document}